\documentclass[11pt]{article}
\usepackage{acl2015}
\usepackage{times}
\usepackage{url}
\usepackage{dsfont}
\usepackage{latexsym}
\usepackage{graphicx}

\title{Document Classification by Inversion of \\Distributed Language Representations}

\author{Matt Taddy \\
  University of Chicago Booth School of Business \\
  {\tt taddy@chicagobooth.edu} \\}

\date{}

\begin{document}
\maketitle
\begin{abstract}
There have been many recent advances in the structure and measurement of {\it distributed} language models: those that map from words to a vector-space that is rich in information about word choice and composition.  This vector-space is the distributed language representation.

The goal of this note is to point out that any distributed  representation can be turned into a classifier through inversion via Bayes rule.  
The approach is simple and modular, in that it will work with any language representation whose training can be formulated as optimizing a probability model. In our application to 2 million sentences from Yelp reviews, we also find that it performs as well as or better than  complex purpose-built algorithms. \end{abstract}

\section{Introduction}

Distributed, or vector-space, language representations $\mathcal{V}$ consist
of a location, or embedding, for every vocabulary {\it word} in $\mathds{R}^K$, where
$K$ is the dimension of the latent representation space.  These locations
are learned to optimize, perhaps approximately, an objective function
defined on the original text such as a likelihood for word occurrences.

A popular example is the Word2Vec machinery of
Mikolov et al.~\shortcite{mikolov_distributed_2013}.  This trains the distributed
representation to be useful as an input layer for prediction of words from
their neighbors in a Skip-gram likelihood.  That is, to maximize
\begin{equation}\label{eq:skipgram}
\sum_{j\neq t,~j=t-b}^{t+b} \log\mathrm{p}_{\mathcal{V}}(w_{sj}\mid w_{st})
\end{equation}
summed across all words $w_{st}$ in all sentences $\mathbf{w}_s$, where $b$ is
the skip-gram window (truncated by the ends of the
sentence) and  $\mathrm{p}_{\mathcal{V}}(w_{sj}| w_{st})$ is a neural network
classifier that takes vector representations for $w_{st}$ and $w_{sj}$
as input (see Section \ref{sec:w2v}).  

Distributed language representations have been studied since the early work on
neural networks \cite{rumelhart_learning_1986} and have long been applied in
natural language processing \cite{morin_hierarchical_2005}.  The models are
generating much recent interest due to the large performance gains from the
newer systems, including Word2Vec and the Glove model of Pennington et
al.~\shortcite{pennington_glove:_2014}, observed in, e.g., word
prediction, word analogy identification, and named entity recognition.

Given the success of these new models, researchers have begun searching for
ways to adapt the representations for use in document classification tasks
such as sentiment prediction or author identification.  One  naive approach is
to use aggregated word vectors across a document (e.g., a document's average
word-vector location) as input to a standard classifier (e.g.,
logistic regression).  However, a document is actually  an {\it ordered} path
of  locations through $\mathds{R}^K$, and simple averaging destroys much of the available
information.  

More sophisticated aggregation is proposed in Socher et al.
\shortcite{socher_parsing_2011,socher_recursive_2013}, where recursive neural
networks are used to combine the word vectors through the estimated parse tree
for each sentence.  Alternatively,  Le and Mikolov's Doc2Vec
\shortcite{le_distributed_2014} adds document labels to the conditioning set
in (\ref{eq:skipgram}) and has them influence the skip-gram likelihood through
a latent input vector location in $\mathcal{V}$. In each case, the end product
is a distributed representation for every sentence (or document for Doc2Vec)
that can be used as input to a generic classifier.

\subsection{Bayesian Inversion}

These approaches all add considerable model and estimation complexity to the
original underlying distributed representation.  We are proposing a
simple alternative that turns fitted distributed language representations into
document classifiers without any additional modeling or estimation.  

A typical language model is trained to maximize the likelihoods of single words and their neighbors.  For example, the  skip-gram
 in (\ref{eq:skipgram}) represents  conditional probability for a
word's context (surrounding words),  while the alternative CBOW Word2Vec
specification \cite{mikolov2013efficient} targets the conditional probability
for each word given its context.  Although these objectives do not correspond to a full document likelihood model, they can be interpreted as components in a \textit{composite likelihood}\footnote{Composite likelihoods are a common tool in analysis of spatial data and data on graphs.  They were popularized in statistics by Besag's \shortcite{besag_spatial_1974,besag1975statistical} work on the pseudolikelihood -- $\mathrm{p}(\mathbf{w}) \approx \prod_j \mathrm{p}(w_j |\mathbf{w}_{-j})$ -- for analysis of Markov random fields. See Varin et al. \shortcite{varin2011overview} for a detailed review.} approximation.

Use $\mathbf{w} = [w_1\dots w_T]'$ to denote a sentence: an ordered vector of words.
The skip-gram  in
(\ref{eq:skipgram}) yields the pairwise composite log likelihood\footnote{See  Molenberghs and Verbeke  \shortcite{molenberghs2006models} for  similar pairwise compositions in analysis of longitudinal data.}
\begin{equation}\label{eq:sentencelhd} \log\mathrm{p}_{ \mathcal{V}}(\mathbf{w}) = 
\sum_{j=1}^T\sum_{k=1}^T \mathds{1}_{\left[1\leq |k-j| \leq b\right]} \log\mathrm{p}_{ \mathcal{V}}(w_{k}|
w_{j} ). \end{equation} 
In another example, Jernite et al.~\shortcite{jernite2015mrf} show that CBOW Word2Vec corresponds to the pseudolikelihood for a Markov random field sentence model.  

Finally, given  a  sentence likelihood as in (\ref{eq:sentencelhd}), document $d =
\{\mathbf{w}_1, ... \mathbf{w}_S\}$ has  log likelihood 
\begin{equation}\label{eq:fulllhd} \log\mathrm{p}_{ \mathcal{V}}(d) = 
\sum_{s}  \log\mathrm{p}_{ \mathcal{V}}(\mathbf{w}_s). \end{equation}

Now suppose that your training documents are grouped by class label, $y \in
\{1 \dots C\}$.  We can train {\it separate} distributed language representations
for each set of documents as partitioned by $y$; for example, fit Word2Vec independently on each sub-corpus $D_c = \{ d_i : y_i =c \}$ and obtain the labeled distributed representation map $\mathcal{V}_c$.  A new  document $d$ has probability
$\mathrm{p}_{ \mathcal{V}_c}(d)$ if we treat it as a member of class $c$, and Bayes rule implies
\begin{equation}\label{eq:bayesrule}
\mathrm{p}( y | d) = \frac{\mathrm{p}_{ \mathcal{V}_y}(d)\pi_y }
{\sum_c \mathrm{p}_{ \mathcal{V}_c}(d)\pi_c }
\end{equation}
where $\pi_c$ is our prior probability on class label $c$.

Thus distributed language representations trained separately for each class label 
yield directly a document classification rule via (\ref{eq:bayesrule}).  This
approach has a number of attractive qualities.

\vskip .1cm
\noindent \textbf{Simplicity:} The inversion strategy works for any model of
language that can (or its training can) be interpreted as a probabilistic
model.  This makes for easy implementation in systems that are already 
engineered to fit such language representations, leading to faster deployment and lower development costs.  
The strategy is also interpretable: whatever intuition one has about the
distributed language model can be applied directly to the 
inversion-based classification rule.  Inversion adds a
plausible model for reader understanding on top of any given language
representation.

\vskip .1cm
\noindent \textbf{Scalability:}  when working with
massive corpora it is often useful to split the data into blocks as part of
distributed computing strategies. Our model of classification via inversion
provides a convenient top-level partitioning of the data.  An efficient system
could fit separate by-class language representations, which
will provide for document classification as in this article as well as
class-specific answers for NLP tasks such as word prediction or analogy.  When
one wishes to treat a document as unlabeled, NLP tasks can be answered through
ensemble aggregation of the class-specific answers.  

\vskip .1cm
\noindent \textbf{Performance:} We find that, in our examples,  inversion of
Word2Vec yields lower misclassification rates than both Doc2Vec-based
classification and the multinomial inverse regression (MNIR) of Taddy
\shortcite{taddy_multinomial_2013}.  We did not anticipate such outright
performance gain.  Moreover, we expect that with calibration (i.e., through
cross-validation) of the  many various tuning parameters available when
fitting both Word and Doc 2Vec the performance results will change.  Indeed,
we find that all methods are often outperformed by phrase-count logistic
regression with rare-feature up-weighting and carefully chosen regularization.
However, the  out-of-the-box performance of Word2Vec inversion
argues for its consideration as a simple default in document classification.

\vskip .2cm
In the remainder, we outline classification through inversion of a specific
Word2Vec model and illustrate  the ideas in classification of Yelp reviews.
The implementation requires only a small extension of the popular
\texttt{gensim} python library \cite{rehurek_software_2010}; the extended
library as well as code to reproduce all of the results in this paper are
available on \texttt{github}. In addition, the yelp data is publicly available
as part of the corresponding data mining contest at
\texttt{kaggle.com}.  
See \texttt{github.com/taddylab/deepir} for detail.

\section{Implementation}
\label{sec:w2v}

\begin{figure}[b]
~\includegraphics[width=0.47\textwidth]{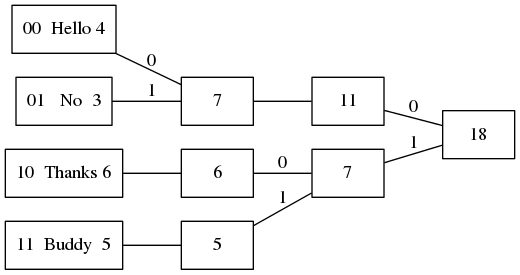}
\caption{\label{bht} Binary Huffman encoding of a 4 word vocabulary, based upon 18 total utterances.  
At each step proceeding from left to right the two nodes with lowest count are
combined into a parent node.  Binary encodings are read back off of the splits
moving from right to left. }
\end{figure}

Word2Vec trains $\mathcal{V}$ to maximize the skip-gram likelihood based on (\ref{eq:skipgram}).  We work with the Huffman softmax specification \cite{mikolov_distributed_2013}, which includes a pre-processing step to encode each vocabulary word in its representation via a binary Huffman tree (see Figure \ref{bht}).

Each individual probability is then
\begin{equation} \label{eq:neuralnet}
\mathrm{p}_{\mathcal{V}}(w | w_t) =\!\!\!
 \prod_{j=1}^{L(w)-1} \!\!\!\sigma\!\left( \mathrm{ch}\left[\eta(w,j+1)\right] \mathbf{u}_{\eta(w,j)}^\top \mathbf{v}_{w_t} \right) 
\end{equation}
where $\eta(w,i)$ is the $i^{th}$ node in the Huffman tree path, of  length $L(w)$, for word $w$; $\sigma(x) = 1/(1 + \exp[-x])$; and $\mathrm{ch}(\eta)
\in \{-1,+1\}$ translates from whether $\eta$ is a left or right child to +/-
1.  Every word thus has both input and output vector coordinates,
$\mathbf{v}_w$ and $[\mathbf{u}_{\eta(w,1)} \cdots \mathbf{u}_{\eta(w,L(w))}]$.
Typically, only the input space $\mathbf{V} = [\mathbf{v}_{w_1} \cdots \mathbf{v}_{w_p}]$,
for a $p$-word vocabulary, is reported as the  language
representation -- these vectors are used as input for NLP tasks.    However,
the full representation $\mathcal{V}$ includes mapping from each word to both
$\mathbf{V}$ and $\mathbf{U}$.

We apply the
\texttt{gensim} python implementation of Word2Vec, which fits the model via stochastic gradient descent (SGD),  under default specification.  This includes a vector space of dimension $K=100$ and a skip-gram window of size $b=5$.  

\subsection{Word2Vec Inversion}

\begin{figure*}

\includegraphics[width=1\textwidth]{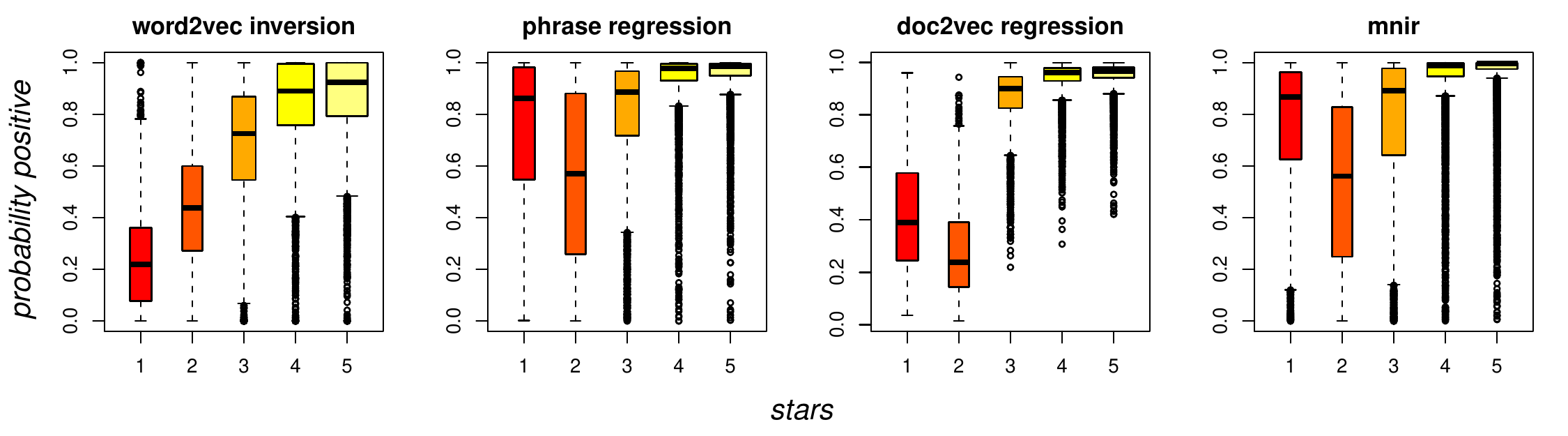}
\vskip -.25cm
\caption{\label{pic:coarseprob} Out-of-Sample fitted probabilities of a review being \emph{positive} (having greater than 2 stars) as a function of the true number of review stars. Box widths are proportional to number of observations in each class; roughly 10\% of reviews have each of 1-3 stars, while 30\% have 4 stars and 40\% have 5 stars.
}
\end{figure*}
 


 
Given Word2Vec trained on each of $C$ class-specific corpora $D_1 \ldots D_C$,
leading to $C$ distinct language representations $\mathcal{V}_1 \dots
\mathcal{V}_C$, classification for new documents is straightforward.  Consider
the $S$-sentence document $d$:  each sentence $\mathbf{w}_s$ is given a 
probability under each representation $\mathcal{V}_c$ by applying the
calculations in (\ref{eq:skipgram}) and (\ref{eq:neuralnet}).  This leads to
the $S \times C$ matrix of sentence probabilities,
$\mathrm{p}_{\mathcal{V}_c}(\mathbf{w}_s)$, and  document probabilities are
obtained 
\begin{equation}
\mathrm{p}_{\mathcal{V}_c}(d) = \frac{1}{S}\sum_s \mathrm{p}_{\mathcal{V}_c}(\mathbf{w}_s).
\end{equation}
Finally, class probabilities are calculated via Bayes rule as in (\ref{eq:bayesrule}).  We use priors $\pi_c = 1/C$, so that classification proceeds by assigning the class
\begin{equation}\label{eq:class}
\hat y = \mathrm{argmax}_c ~~\mathrm{p}_{\mathcal{V}_c}(d).
\end{equation}

\section{Illustration}

We consider a corpus of reviews provided by Yelp for a contest on {\tt
kaggle.com}.  The text is tokenized simply by converting to lowercase before splitting on punctuation and white-space.  The training data are 230,000 reviews containing more than 2
million sentences. Each review is marked by a number of {\it stars}, from 1
to 5, and we fit separate Word2Vec representations $\mathcal{V}_1 \ldots
\mathcal{V}_5$ for the documents at each star rating.  The validation data
consist of 23,000 reviews, and we apply the inversion technique of Section
\ref{sec:w2v} to score each validation document $d$ with class probabilities
$\mathbf{q} = [q_1 \cdots q_5]$, where $q_c = \mathrm{p}(c|d)$.

The probabilities will be used in three different classification tasks; for reviews as 

\vskip .1cm
$a.$ negative at 1-2 stars, or positive at 3-5 stars; 

\vskip .1cm
$b.$ negative 1-2, neutral 3, or positive 4-5 stars;

\vskip .1cm
$c.$ corresponding to each of 1 to 5 stars.

\vskip .1cm
In each case, classification proceeds by summing across the relevant
sub-class probabilities.  For example, in task $a$,
$\mathrm{p}(\texttt{positive}) = q_3+q_4+q_5$. Note that the same five fitted
Word2Vec representations are used for each task.

We consider a set of related comparator techniques.  In each case, some
document representation (e.g., phrase counts or Doc2Vec vectors) is used as
input to logistic regression prediction of the associated review rating.
The logistic regressions are fit under $L_1$ regularization with the
penalties weighted by feature standard deviation (which, e.g., up-weights rare
phrases) and selected according to the corrected AICc criteria
\cite{flynn_efficiency_2013} via the \texttt{gamlr} R package of Taddy
\shortcite{taddy_one-step_2014}.  For multi-class tasks $b$-$c$, we use
distributed Multinomial regression (DMR; Taddy
2015)\nocite{taddy_distributed_2015} via the \texttt{distrom} R package.  DMR
fits multinomial logistic regression in a factorized representation wherein
one estimates independent Poisson linear models for each response category.
Document representations and logistic regressions are
always trained using only the training corpus.

\vskip .1cm
\noindent \textit{Doc2Vec} is also fit via \texttt{gensim}, using the same
latent space specification as for Word2Vec: $K=100$ and $b=5$.  
As recommended in the documentation, we apply repeated SGD over 20 re-orderings of each
 corpus (for comparability, this was also done when fitting Word2Vec).
Le and Mikolov provide two alternative Doc2Vec specifications: distributed
memory (DM) and distributed bag-of-words (DBOW).  We fit both. Vector representations for validation documents are trained without
updating the word-vector elements, leading to 100 dimensional vectors for
each document for each of DM and DCBOW.  We input each, as well as the combined 200 dimensional
DM+DBOW representation, to logistic regression.

\vskip .1cm
\noindent \textit{Phrase regression} applies logistic regression of 
response classes directly onto counts for short 1-2 word `phrases'.  The phrases are
obtained using \texttt{gensim}'s phrase builder, which simply combines highly
probable pairings; e.g., \texttt{first\_date} and
\texttt{chicken\_wing} are two pairings in this corpus.  

\vskip .1cm
\noindent \textit{MNIR}, the multinomial inverse regression of Taddy
\shortcite{taddy_measuring_2013,taddy_multinomial_2013,taddy_distributed_2015}
is applied  as implemented in the \texttt{textir} package for R.  MNIR maps
from text to the class-space of interest through a multinomial logistic
regression of phrase counts onto variables relevant to the class-space. We
apply MNIR to the same set of 1-2 word phrases used in phrase regression.
Here, we regress phrase counts onto stars expressed numerically and as a
5-dimensional indicator vector, leading to a 6-feature multinomial logistic
regression.  The MNIR procedure then uses the $6\times p$ matrix of 
feature-phrase regression coefficients to map from phrase-count to feature space,
resulting in 6 dimensional `sufficient reduction' statistics for each
document.  These  are input to logistic
regression.

\vskip .1cm
\noindent \textit{Word2Vec aggregation}  averages  fitted word
representations for a single Word2Vec trained on all sentences to obtain a
fixed-length feature vector for each review ($K=100$, as for inversion).  This
vector is then input to logistic regression.


\begin{table}
\hspace{-.25cm}
{
\begin{tabular}{r|c c c}
& $a$ (NP) & $b$ (NNP) &  $c$ (1-5)
\\ \cline{2-4}\rule{0pt}{3ex}
W2V inversion & .099 & \textbf{.189} & .435 \\
Phrase regression & \textbf{.084} & .200 & \textbf{.410} \\
D2V DBOW &  .144 &.282 & .496 \\
D2V DM & .179 & .306 & .549 \\
D2V combined & .148 & . 284 & .500 \\
MNIR & .095 & .254 & .480 \\
W2V aggregation & .118 & .248 & .461 
\end{tabular}}
\caption{ Out-of-sample misclassification rates.}
\end{table}

\subsection{Results}

Misclassification rates for each task on the validation set are reported in
Table 1. Simple phrase-count regression is consistently the
strongest performer, bested only by Word2Vec inversion on task $b$.  This is
partially due to the relative strengths of discriminative (e.g., logistic
regression) vs generative (e.g., all others here) classifiers: given a large amount of
training text, asymptotic efficiency of logistic regression will start to work
in its favor over the finite sample advantages of a generative classifier
\cite{ng_discriminative_2002,taddy_rejoinder:_2013}.
However, the comparison is also unfair to Word2Vec and Doc2Vec: both
phrase regression and MNIR are optimized exactly under
AICc selected penalty, while Word and Doc 2Vec have only been approximately
optimized under a single specification.  The
distributed representations should improve  with some careful engineering.

Word2Vec inversion outperforms the other document representation-based
alternatives (except, by a narrow margin, MNIR in task $a$).  Doc2Vec under
DBOW specification and MNIR both do worse, but not by a large margin. In
contrast to  Le and Mikolov, we find here that the Doc2Vec DM model does much
worse than DBOW.  Regression onto simple within- document aggregations of
Word2Vec perform slightly better than any Doc2Vec option (but not as well as
the Word2Vec inversion).  This again contrasts the results of Le and Mikolov
and we suspect that the more complex
Doc2Vec model would benefit from a careful tuning of the SGD optimization
routine.\footnote{Note also that the unsupervised document representations -- Doc2Vec or the single Word2Vec used in Word2Vec aggregation -- could be trained on larger unlabeled corpora.  A similar option is available for Word2Vec inversion: one could take a single Word2Vec model trained on a large unlabeled corpora as a shared baseline (prior) and  update separate models with additional training on each labeled sub-corpora.  The representations will all be shrunk towards a baseline language model, but will differ according to distinctions between the language in each labeled sub-corpora.}

Looking at the fitted probabilities in detail we see that Word2Vec inversion
provides a more useful document {\it ranking} than any comparator (including
phrase regression).  For example, Figure \ref{pic:coarseprob} shows the
probabilities of a review being `positive' in task $a$ as a function of the
true star rating for each validation review. Although phrase regression does
slightly better in terms of misclassification rate, it does so at the cost of
classifying many terrible (1 star) reviews as positive.  This occurs  because 1-2 star reviews are more rare than 3-5 star reviews and because words of emphasis (e.g. \texttt{very}, \texttt{completely}, and \texttt{!!!}) are used both in very bad and in very good reviews.  Word2Vec inversion is
the {\it only} method that yields positive-document probabilities that are
clearly increasing in distribution with the true star rating.  It is not
difficult to envision a misclassification cost structure that favors such
nicely ordered probabilities.

\section{Discussion}

The goal of this note is to point out inversion as an option for turning distributed language representations into classification rules.  We are not arguing for the supremacy of Word2Vec inversion in particular, and the approach should work well with alternative representations (e.g., Glove).  Moreover, we are not even arguing that it will always outperform purpose-built classification tools.  However, it is a simple, scalable, interpretable, and effective option for classification whenever you are working with such distributed representations.

\bibliographystyle{acl}
\bibliography{deepir}

\end{document}